\documentclass[sn-mathphys-ay]{sn-jnl1}

\usepackage{graphicx}
\usepackage{multirow}
\usepackage{amsmath,amssymb,amsfonts}
\usepackage{amsthm}
\usepackage{mathrsfs}
\usepackage[title]{appendix}
\usepackage{textcomp}
\usepackage{manyfoot}
\usepackage{booktabs}
\usepackage{float}
\usepackage{caption}
\usepackage{algorithm}
\usepackage{algorithmicx}
\usepackage{algpseudocode}
\usepackage{listings}
\usepackage{subcaption}
\usepackage{array}
\usepackage{tabularx}
\usepackage[utf8]{inputenc}
\usepackage{siunitx}
\usepackage[authoryear]{natbib}

\theoremstyle{plain}
\theoremstyle{thmstyleone}

\theoremstyle{thmstyletwo}

\theoremstyle{thmstylethree}

\usepackage{hyperref}

\begin{document}

\title[Article Title]{Odin: Oriented Dual-module Integration for Text-rich Network Representation Learning}

\author[1]{\fnm{Kaifeng} \sur{Hong}}
\author*[1]{\fnm{Yinglong} \sur{Zhang}}\email{zhang\_yinglong@126.com}
\author[1]{\fnm{Xiaoying} \sur{Hong}}

\author[1]{\fnm{Xuewen} \sur{xia}}

\author[1]{\fnm{Xing} \sur{Xu}}

\affil[1]{\orgdiv{College of Physics and Information Engineering}, \orgname{Minnan Normal University}, \orgaddress{\street{Zhangzhou}, \city{Fujian}, \postcode{363000}, \country{China}}}


\abstract{Text-attributed graphs require models to effectively combine strong textual understanding with structurally informed reasoning. Existing approaches either rely on GNNs—limited by over-smoothing and hop-dependent diffusion—or employ Transformers that overlook graph topology and treat nodes as isolated sequences. We propose \textbf{Odin} (\textbf{O}riented \textbf{D}ual-module \textbf{IN}tegration), a new architecture that injects graph structure into Transformers at selected depths through an oriented dual-module mechanism. Unlike message-passing GNNs, Odin does not rely on multi-hop diffusion; instead, multi-hop structures is integrated at specific Transformer layers, yielding low-, mid-, and high-level structural abstraction aligned with the model’s semantic hierarchy. Because aggregation operates on the global [CLS] representation, Odin fundamentally avoids over-smoothing and decouples structural abstraction from neighborhood size or graph topology. We further establish that Odin’s expressive power strictly contains that of both pure Transformers and GNNs. 
	To make the design efficient in large-scale or low-resource settings, we introduce Light Odin, a lightweight variant that preserves the same layer-aligned structural abstraction for faster training and inference. Experiments on multiple text-rich graph benchmarks show that Odin achieves state-of-the-art accuracy, while Light Odin delivers competitive performance with significantly reduced computational cost. Together, Odin and Light Odin form a unified, hop-free framework for principled structure–text integration. The source code of this model has been released at https://github.com/hongkaifeng/Odin. 
}

\keywords{}



\maketitle

\section{Introduction}\label{sec1}

Text-Attributed Graphs (TAGs) \citep{ref1,ref3,ref12,ref13,ref24}deeply integrate node text semantics and graph topology. They combine textual descriptions of entities (e.g., paper abstracts, product descriptions, user reviews) with relationships between entities (e.g., citation links, product purchase connections, social friend interactions), and have become the core data form in fields such as information retrieval \citep{ref1,ref5}, product recommendation \citep{ref11}, academic network analysis \citep{ref4}, and document recommendation \citep{ref5}.

Single-modal modeling of such data has obvious limitations \citep{ref24}. Using only graph structure can capture topological connections but fails to achieve fine-grained understanding of entity semantics—for instance, in academic citation networks, it is hard to distinguish between different types of citations (those on similar topics or those for method comparison); using only text information can parse semantic connotations but overlooks potential connections—for example, papers on similar topics may be misjudged due to the absence of direct citations. In recent years, research on dual-modal fusion for text-attributed graphs based on Language Models (LMs) and Graph Neural Networks (GNNs) \citep{ref1,ref2,ref37,ref38,ref42,ref43}has remedied the shortcomings of single-modal approaches and provided more comprehensive feature support for downstream tasks.

Current mainstream methods for modeling based on LMs and GNNs can be categorized into three categories \citep{ref3}: LLM-as-Enhancer \citep{ref37,ref38}, LLM-as-Predictor  \citep{ref2,ref42}, and LLM-as-Aligner \citep{ref1,ref43}. In the LLM-as-Enhancer paradigm, large language models are employed to interpret or refine textual semantics, thereby enriching node features\citep{ref3}. The LLM-as-Predictor paradigm reformulates graph learning tasks as text generation or sequence prediction problems. In contrast, the LLM-as-Aligner paradigm seeks to jointly model textual semantics and graph structure within a unified architecture. 

Despite significant progress in current research, studies based on LMs and GNNs are still in their infancy, facing several fundamental challenges. The LLM-as-Enhancer method suffers from insufficient interaction between semantics and structure. The LLM-as-Predictor is prone to losing high-order structural information and has weak output controllability. Although LLM-as-Aligner approaches enable joint processing of text semantics and graph structure, they suffer from issues such as the substantial layer-depth mismatch between GNNs and Pre-trained language models (PLMs), the problem of rigid structure injection, and a persistent trade-off between efficiency and performance. Motivated by these limitations, this work develops a new solution within the LLM-as-Aligner paradigm.

GNNs are effective at extracting low-, mid-, and high-order structural information from graph topology via neighborhood aggregation. PLMs, in contrast, refine semantic representations layer by layer through Transformers: shallow layers capture lexical associations and local syntactic cues, middle layers integrate phrase-level and contextual semantics, and deeper layers distill global, abstract meaning. However, mainstream GNNs typically use only 2–3 layers to avoid over-smoothing, whereas Transformers often require 12 or more layers to capture deep semantics. This mismatch—structural depth (M) is far smaller than semantic depth (L)—creates a fundamental barrier to deep fusion. Moreover, increasing the number of GNN layers exacerbates neighborhood expansion: deeper message passing enlarges the computational graph, and under the parallel processing paradigm of Transformers, this leads to a rapid growth in the number of texts that must be processed simultaneously, dramatically increasing memory and computation cost.

GraphFormers \citep{ref1} and Patton \citep{ref24}, representative models based on LLM-as-Aligner, also suffer from an imbalance between structure and semantic adaptation and fail to achieve phased and precise fusion. Although these two types of models insert GNN modules between Transformer layers to realize the fusion of GNNs and Transformers, all Transformer layers share the same set of GNN parameters, leading to the problem of rigid structure injection. On the other hand, to reduce computational resource overhead, these models only deploy a single GNN layer, whose perception range is limited to the 1-hop neighborhood of the central node. As a result, they can only extract low-order structural information and lack the ability to capture high-order neighborhood structural information, making it difficult to match the abstract semantic associations required by the deep Transformer network. Furthermore, in large-scale networks, high-order neighbors may contain important information. 

Taking figure 1 as an example, the meaning of ``Rose" cannot be determined from the central node alone; it may refer either to the flower or to a person named Rose. Incorporating 1-hop neighbors—such as flower and goods—increases the likelihood that ``Rose" denotes the flower, yet this evidence remains insufficient to exclude the possibility of a person (e.g., a local notable figure). However, after integrating 2-hop neighbors—including herbs, roses, and rare roses—the broader contextual information strongly indicates that ``Rose" refers to the flower rather than a person. This example illustrates a fundamental principle of text-rich graph modeling: multi-hop structural context can serve as valuable semantic evidence for language models, enabling them to resolve lexical ambiguities and construct more accurate representations. However, prior LLM-as-Aligner models typically rely on a single GNN layer, restricting their receptive field to only 1-hop neighbors. As a result, they capture merely low-order structural information and fail to supply the deeper contextual cues required by the upper layers of a Transformer.

\begin{figure}[h] 
	\centering
	\includegraphics[width=0.9\textwidth]{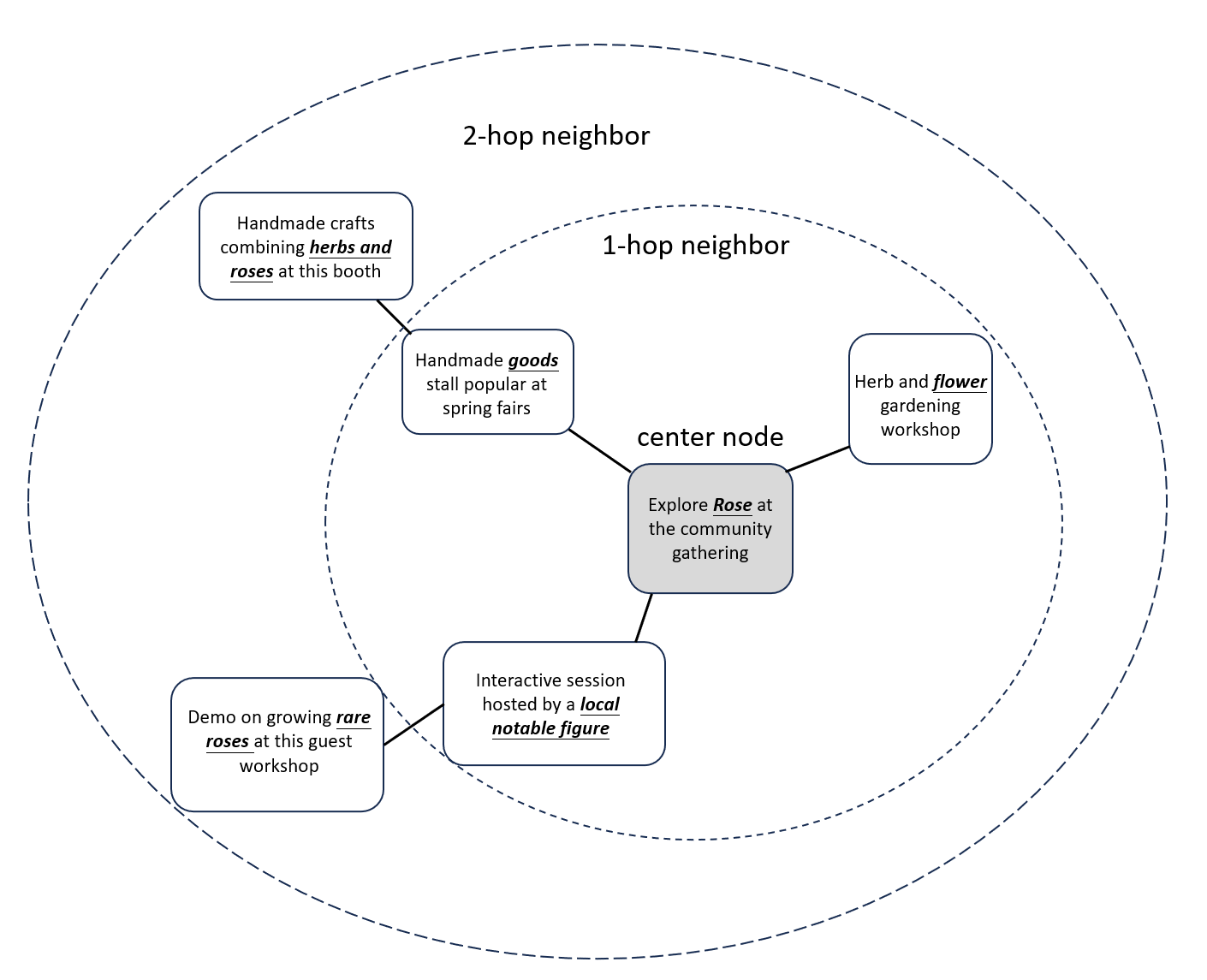}  %
	\caption{An illustration of a text-rich network where multi-hop neighborhood information provides crucial contextual cues for semantic interpretation.} 
	\label{fig:citation_network} 
\end{figure}

To address the above challenges, we propose Odin, a framework founded on layer-aligned structural fusion and lightweight adaptation. Odin is driven by a central insight: although multi-hop structural information inherently contains low-, mid-, and high-order cues, the resulting low-, mid-, and high-level structural abstractions should not be dictated by neighborhood size. Instead, these abstractions should emerge from the specific positions of structure-injection layers within the Transformer hierarchy. In other words, structural information extracted from multi-hop neighborhoods should be injected at the semantic depths where it is most compatible—low-order structural cues aligned with shallow semantic layers, mid-order structural patterns aligned with intermediate semantic reasoning, and high-order or global structure aligned with the deep semantic layers of the Transformer. To operationalize this principle, Odin employs a hierarchical dual-module design. GNN aggregation layers with stage-specific parameters  (removing parameter-sharing constraints) are inserted into the Transformer’s semantically critical layers, enabling precise and depth-aligned structural fusion. Simple structural aggregation layers are placed in non-critical layers to maintain continuous structural conditioning through simple aggregation, ensuring full-stage guidance while avoiding unnecessary computational expansion.
For large-scale applications, we further develop Light Odin, a compact variant that preserves Odin’s layer-aligned structural abstraction using only a 6-layer Transformer, substantially reducing computational overhead with only minor performance degradation. Light Odin achieves an effective balance between efficiency and performance in text-attributed graph scenarios.
In summary, our main contributions are as follows:

\begin{enumerate}
	\item We propose Odin, a framework for phased and depth-aligned fusion of structural and semantic signals. Odin resolves the long-standing layer-mismatch between PLMs and GNNs by establishing a one-to-one alignment between the differentiated, stage-specific GNN aggregation layers and the Transformer’s semantic hierarchy. Together with simple structure aggregation layers inserted into non-critical depths, Odin enables—for the first time—full-stage adaptation and interactive enhancement between low-, mid-, and high-level structural abstractions and the corresponding semantic layers of the Transformer.
	\item We develop Light Odin, a compact variant that preserves structure–semantics alignment with significantly improved efficiency. While maintaining over 85\% of the original performance, it reduces the parameter scale and inference overhead by 60\%, addressing the efficiency bottleneck of large-scale text-attributed graphs.
	\item Extensive experiments on five real-world datasets demonstrate the effectiveness of our approach. Odin consistently outperforms state-of-the-art baselines such as LLAGA and PATTON across most benchmarks, while Light Odin achieves a favorable balance between accuracy and computational efficiency, highlighting its strong applicability in real-world scenarios.
\end{enumerate}

\section{Related Work}
\subsection{PLMs}
Since BERT \citep{ref17} proposed the bidirectional Transformer architecture, PLMs have entered a period of rapid iteration. EMLo \citep{ref30} first integrated contextual information into word embeddings; GPT \citep{ref31} realized generative language modeling via unidirectional Transformers; XLNet \citep{ref32} introduced permutation language modeling to break through BERT's bidirectional semantic limitations; T5 \citep{ref33} unified various NLP tasks under the ``text-to-text" paradigm; GPT-3 \citep{ref34} achieved generalization in zero-shot scenarios through a 100-billion-scale parameter size. Such models can generate node-level representations containing contextual semantics (e.g., BERT's [CLS] token) via pre-training tasks such as Masked Language Modeling (MLM) and Next Sentence Prediction (NSP). As advanced representatives of current general text semantic modeling, ModernBERT \citep{ref25} further optimizes the deep architecture design and adopts a ``dynamic positional encoding" mechanism — abandoning traditional fixed positional vectors and adaptively adjusting positional encoding weights with the length of text sequences to effectively alleviate semantic bias in long texts (e.g., academic paper abstracts, detailed product descriptions); additionally, it performs better in modeling long node texts in text-attributed graphs. SciBERT \citep{ref23} is pre-trained on 1.14 million academic papers to enhance the semantic capture of academic terms.
Targeting the domain characteristics of text-attributed graphs, LinkBERT \citep{ref26} learns cross-document semantic associations through the document link prediction task. These PLMs have shown highly competitive performance in graph tasks.

\subsection{GNN}

GNNs are core tools for structural modeling of text-attributed graphs, converting topology into node embeddings through message passing. Their development can be divided into two categories: Classical transductive GNNs: Represented by GCN \citep{ref35} and GAT \citep{ref36}, GCN relies on the global Laplacian matrix (transductive limitation) and is prone to over-smoothing; GAT dynamically assigns neighborhood weights using attention. Both rely only on simple text feature initialization, with the number of layers limited to 2~3, making it difficult to match the deep semantic needs of PLMs. Inductive GNNs and their variants: GraphSAGE \citep{ref16} supports inductive learning through ``neighborhood sampling and feature aggregation"; GraphSAGE++ \citep{ref27} introduces weighted multi-scale aggregation on the basis of GraphSAGE to improve structural discriminability. GNNs have limitations: their performance depends on the quality of initial node attributes, and structure is disconnected from text semantics.

\subsection{Fusion Paradigms of Large Language Models and Graph Neural Networks}

In recent years, with the rise of Large Language Models (LLMs), researchers have begun to focus on how to introduce the strong semantic modeling capabilities of LLMs into graph tasks, forming an emerging research direction of ``Graph\&LLM". According to the functional role of LLMs in the framework, current mainstream methods can be divided into three categories: LLM-as-Enhancer, LLM-as-Predictor, and LLM-as-Aligner.

\subsubsection{LLM-as-Enhancer}

This type of method uses LLMs as feature enhancers to improve the semantic quality of node embeddings using LLMs before structural aggregation. Typical representatives include GIANT \citep{ref38}, TAPE \citep{ref37}, GraphPro \citep{ref46}, SKETCH \citep{ref43}, OFA \citep{ref39}, and ENGINE \citep{ref40}. GIANT injects structural signals into the BERT representation space through a neighborhood prediction self-supervised task; TAPE enhances node semantics using explanatory text generated by GPT-3.5 or LLaMA; OFA introduces a Graph Prompt mechanism to achieve cross-task transfer; ENGINE realizes low-cost structural enhancement by adding lightweight graph convolution layers (G-Ladder) next to the frozen LLM backbone; SKETCH (ACL 2025) decouples node aggregation from graph convolution and places it before the LM training stage, achieving more efficient TAGs learning with fewer computational resources without relying on GNNs; STAG (KDD 2025) addresses the core pain points of text-attributed graphs (TAGs) where structural information is difficult to convert into LLM-compatible formats and cross-domain transfer relies on source domain labels, proposing a soft tokenization framework (STAG). GraphPro \citep{ref46} focuses on dynamic recommendation TAGs (user-item interaction graphs with text descriptions for nodes), inputting the enhanced semantic embeddings into GNNs for multi-hop structural aggregation to adapt to TAG tasks in dynamic scenarios. Such methods have simple structures and high compatibility, but their cascaded logic of ``semantic enhancement → structural aggregation" keeps LLMs and GNNs independent of each other, making it difficult to achieve multi-level information interaction within the model.
The Odin framework further deepens on the basis of such methods: by alternately inserting GNN aggregation layers and simple graph structure aggregation layers into each layer of the Transformer, structural information continuously participates in different stages of semantic modeling, thereby realizing true hierarchical semantic-structural co-construction. This mechanism breaks through the limitations of shallow structure injection and discontinuous fusion in previous LLM-as-Enhancer models.

\subsubsection{LLM-as-Predictor}

This type of method converts graph tasks into natural language tasks, with LLMs directly completing prediction or generation. Representative works include GraphText \citep{ref41}, GraphAdapter \citep{ref42}, LLaGA \citep{ref2}, STAG \citep{ref14}, etc. They usually design textual prompts for graph structures, convert node attributes, neighbor information, and task descriptions into language inputs, and then LLMs output node classification or edge prediction results. STAG encodes graph structures through GNNs, combines the output with system prompts, and inputs them into LLMs to generate category prediction text. This type of method has advantages such as strong interpretability and flexible transfer, and naturally supports zero-shot inference. However, due to the easy loss of high-order structural information during the graph-to-text conversion process and weak controllability of generated outputs, their performance fluctuates greatly on complex structural data (e.g., citation networks, knowledge graphs). In contrast, Odin does not fully rely on generative LLMs; instead, it gradually injects structural context into the multi-layer semantic space of the Transformer through an internal hierarchical alignment mechanism, enabling the model to have cross-level semantic enhancement capabilities while maintaining controllability.

\subsubsection{LLM-as-Aligner}

This paradigm achieves deep synergy between LLMs and GNNs through joint optimization or spatial alignment mechanisms, with representative works including GLEM \citep{ref12}, GraphFormer \citep{ref1}, Patton \citep{ref24}, SSTAG \citep{ref47}, and GraphCLIP \citep{ref49}. GLEM adopts an EM framework to alternately optimize LLMs and GNNs, enabling dynamic complementarity between the two in the embedding space; GraphFormer embeds GNN modules between Transformer layers to achieve structure-semantic fusion; Patton designs network context pre-training tasks (NMLM and MNP) to realize bidirectional enhancement of semantic and structural layers. SSTAG designs a dual knowledge distillation and memory anchor alignment framework: first, it extracts text semantic embeddings through LLMs and structural embeddings through GCNs, then uses student-teacher consistency loss to make lightweight MLPs fit the two types of features. GraphCLIP targets cross-modal alignment tasks for TAGs, using LLMs as aligners for text-structural features. This type of method has the greatest potential in terms of fusion depth, but has two prominent problems: (1) Depth Mismatch — GNNs have shallow structures while Transformers are deep, leading to unsynchronized semantic and structural fusion; (2) Rigid structure injection and over-smoothing risk — excessive accumulation of graph information in deep layers will destroy the hierarchical feature abstraction of Transformers.
Odin is a new-generation deep fusion solution proposed to address this type of problem. It realizes phased alignment between low-, middle-, and high-level semantic layers and corresponding structural abstractions, thereby maintaining structural guidance in all stages of semantic abstraction.

\section{Preliminaries}

\subsection{Basic Definition}

Let $\mathcal{G} = (\mathcal{V}, \mathcal{E}, \mathbf{X})$ be a \textbf{text-attributed graph (TAG)}, where $\mathcal{V}$ is the set of nodes, $\mathcal{E} \subseteq \mathcal{V} \times \mathcal{V}$ is the set of edges encoding relational structure, $\mathbf{X} = \{x_v \mid v \in \mathcal{V}\}$ denotes the raw text associated with each node. The goal is to learn a node representation $\mathbf{h}_v \in \mathbb{R}^d$ that jointly captures (1) its \textbf{text semantics} and (2) its \textbf{multi-hop structural context}.

\subsection{Introduction to GNNs}

Graph Neural Networks (GNNs) are a class of deep learning models specifically designed to process graph-structured data. Their core advantage lies in converting topological associations of graphs into quantifiable node embeddings by simulating the Message Passing mechanism between nodes, thereby effectively capturing local and global structural features of graph data. Among numerous GNN models, GraphSAGE \citep{ref16} is a classic framework for inductive GNNs. Its core breakthrough is solving the limitation of traditional graph representation learning methods that ``cannot handle unseen nodes". Through the modular design of ``neighborhood sampling and feature aggregation", it realizes efficient inductive node embedding generation and becomes one of the mainstream tools for structural modeling of text-attributed graphs. In the aggregation stage, GraphSAGE aggregates information from node neighborhoods through an aggregation function. Taking the mean function as an example of the aggregation function, the mathematical formula of the aggregation function is:\begin{equation*}h_v^{(k)} = \sigma\left( W \cdot \text{MEAN}\left( { h_v^{(k-1)} } \cup { h_u^{(k-1)} \mid u \in \mathcal{N}(v) } \right) \right)\end{equation*}where \(h_v^{(k)}\) denotes the representation of node v at layer k, \(\sigma\) is a non-linear activation function, and W is a set of trainable weights. \(\text{MEAN}(\cdot)\) denotes the mean calculation operation, whose input includes the embedding \(h_v^{(k-1)}\) of node v at the previous layer and the embeddings \(h_u^{(k-1)}\) of all nodes u in the neighborhood set \(\mathcal{N}(v)\) of node v at the previous layer.

\subsection{Transformer-Based Language Models}

BERT \citep{ref17}, a pre-trained language model proposed by Devlin et al. (2018), achieves its core breakthrough by capturing deep text contextual semantics through a bidirectional Transformer architecture and self-supervised pre-training tasks, providing a powerful semantic encoding tool for natural language processing (NLP) and node text modeling of text-attributed graphs. BERT is composed of 12 stacked Transformer encoder blocks with the same structure. Each Transformer block realizes the layer-by-layer abstraction and deepening of text semantics through a series structure of self-attention mechanism and multi-layer perceptron. The following operations are performed in each Transformer block:

$$
		h^{(l)'} = \text{LN}\left( h^{(l-1)} + \operatorname{AsymmetricMultiHead}\left( h^{(l-1)} \right) \right) \\
$$
$$
		h^{(l+1)} = \text{LN}\left( h^{(l)'} + \text{MLP}\left( h^{(l)'} \right) \right)
$$
where $h^{(l)}$ denotes the hidden state of the text at layer $l$, $\operatorname{AsymmetricMultiHead}(\cdot)$ is asymmetric multi-head self-attention, $\text{LN}(\cdot)$ is the layer normalization operation, and $\text{MLP}(\cdot)$ is the multi-layer perceptron.

\subsection{Differences Between Transformer-Based Language Models and GNNs}

To clarify the functional positioning of BERT and GNNs in text-attributed graph modeling, this section takes GraphSAGE and BERT-base as examples and presents their core differences from four aspects (data adaptation, mechanism design, performance advantages, and architecture configuration) in the form of a table, as shown in Table 1.

\begin{table}[htbp]
	\centering
	\caption{Differences Between GNNs and BERT}
	\label{tab:gnn_bert_diff} 
	\begin{tabularx}{\textwidth}{lXXXX}
		\toprule
		Model & Data Processed & Core Mechanism & Advantages & Number of Layers \\
		\midrule
		GNN & Graph-Structured Data & Message Passing Mechanism & Capture complex relationships and dependencies between nodes & 2 \\
		BERT & 1D Text Sequences & Self-Attention Mechanism & Deeply understand contextual relationships of text & 12 \\
		\bottomrule
	\end{tabularx}
\end{table}

\section{Odin}

Text-attributed graphs encode multi-hop structural signals that are essential for resolving semantic ambiguities, yet Transformers overlook these non-Euclidean dependencies and existing LLM-as-Aligner models capture only shallow (1-hop) structure, limiting their ability to align with deep semantic layers. This motivates Odin, which rethinks structure–semantics fusion by extracting low-, mid-, and high-level structural abstractions from multi-hop neighborhoods and integrating them at carefully selected Transformer layers to achieve depth-aligned, hop-free structural adaptation.

\subsection{Model Framework}
To resolve the fundamental layer-mismatch challenge, Odin introduces a hierarchical alternating design that strategically interleaves TG layers (Text–Graph Fusion layers) and TS layers (Text Encoding and Simple Aggregation layers). This design enables the model to process graph structure as contextual information at Transformer depths where the corresponding semantic complexity naturally aligns with the structural abstraction level.

The core architecture of Odin is shown in Figure 2. The central innovation lies in Odin’s layer-aligned alternating strategy between TG and TS layers—the first architecture to directly reconcile the contradiction between few GNN layers and many PLM layers. By deploying TG layers in key semantic stages to achieve deep fusion and using TS layers for lightweight guidance in non-key stages, it breaks through the limitation of unbalanced structural information injection (premature dilution or late ineffectiveness) in traditional fusion models, realizing dynamic and precise adaptation of structural information in all semantic stages of the Transformer.

For the subgraph obtained after A-hop GraphSAGE mini-batch sampling [16] on the TAG, the text of nodes in the subgraph is divided into token sequences using a tokenizer, and the initial embedding \(T_v^0\) for node text is generated using BERT's embedding initialization. The hidden state of the [cls] token is used as the overall representation of the corresponding node, denoted as \(T_v^0[\text{cls}]\). Odin consists of \textit{L} layers, including \textit{M} TG layers and the remaining TS layers (\textit{L} depends on the number of layers of the pre-trained language model used, \textit{M} depends on the number of layers of the GNN used, and \(L>M\)). TG layers include a text encoding module and a graph aggregation module, while TS layers include a text encoding module and a simple aggregation module. In the TG layer, the text encoder processes to obtain \(T_v^l\). Then, \(T_v^l[\text{cls}]\) is extracted as the node-level representation in the graph, and GNN convolution is used to aggregate neighbor information to generate the node-level enhanced token \(agg_v\). As neighbor context, \(agg_v\) is concatenated with \(T_v^l\) before entering the next layer. The processing flow in the TS layer is roughly the same as that in the TG layer. The difference is that in the TS layer, \(agg_v\) is generated using a simple aggregation method instead of GNN.

\begin{figure}[h] 
	\centering
	\includegraphics[width=0.9\textwidth]{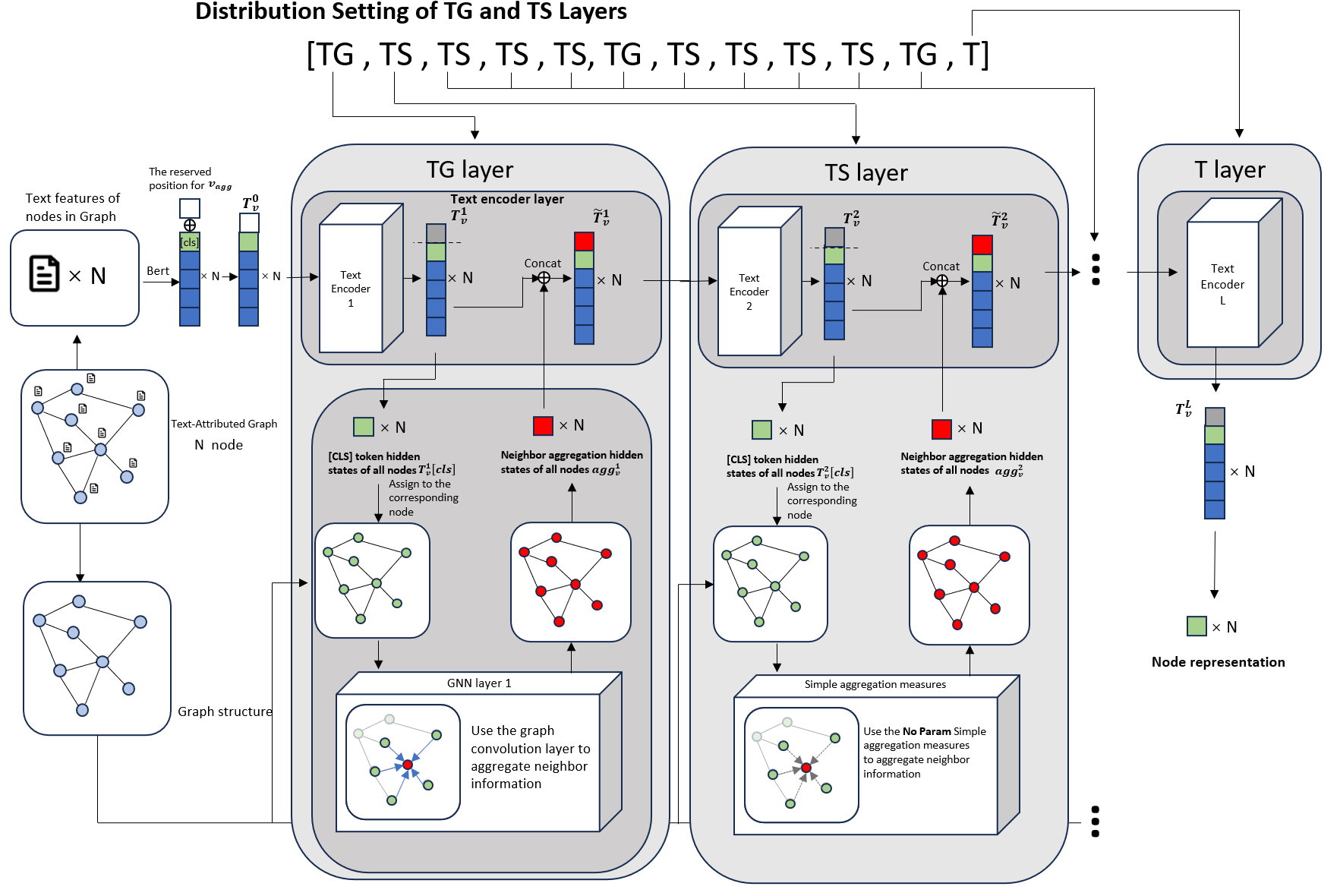}  %
	\caption{Odin Model Structure Diagram} 
	\label{fig:citation_network} 
\end{figure}

\subsection{Graph Aggregation in Odin}

In Odin, with the exception of the final layer, each dual-module integration layer performs either graph aggregation or simple aggregation immediately after its text encoding module. 

The graph aggregation module is the core component of the TG layer. Its main function is to convert the neighborhood structural information of nodes into the semantic context of text encoding, enabling the subsequent text encoding module to simultaneously perceive the topological associations of nodes during modeling and avoiding text understanding from being detached from structural constraints. 

For TS layers without a graph aggregation module, Odin uses a simple aggregation method after the text encoding module of these layers. The core goal of introducing the simple aggregation method is to retain the guidance of structural information on text encoding without adding additional graph aggregation modules (improving accuracy and avoiding the rise in computational complexity), and prevent the text encoding module from regressing to the limitation of ``independent text modeling". The specific formula is:
\begin{equation}
	\text{agg}_v^{(l)} =\begin{cases}W_1^{(l)} \cdot \text{mean}\left( { T_j^{(l)}[\text{cls}] \mid j \in \mathcal{N}(v) } \right) + W_2^{(l)} \cdot T_v^{(l)}, & l \in S \quad (1a) \\{SimpleAggregate}\left( T_v^{(l)}[\text{cls}], { T_j^{(l)}[\text{cls}] \mid j \in \mathcal{N}(v) } \right), & \text{else} \quad (1b)\end{cases}\label{eq:graph_aggregation}
\end{equation}
where \(T_v^{(l)}[\text{cls}]\) is the node-level representation of node v at layer \textit{l}, \(W_1^{(l)}\) and \(W_2^{(l)}\) are the weight parameters of the graph aggregation module at layer \textit{l}, \(\text{mean}(\cdot)\) denotes the mean operation, \(\mathcal{N}(v)\) denotes the neighborhood of node v, and \textit{S} is a list storing the positions of TG layers with length M (M is the total number of TG layers in Odin), where each element represents the position of a TG layer in Odin. Taking \(S=[1,6,11]\) as an example, the 1st, 6th, and 11th layers of Odin are TG layers at this time, and the rest are TS layers. \(\text{SimpleAggregate}(\cdot)\) is the optional simple aggregation strategy used by TS layers, and four simple aggregation strategies and their detailed operations are provided in Section 4.4.\(\text{agg}_v^{(l)}\) enhances the ability of text encoding to perceive graph structures and improves the text encoding performance. The specific operation is as follows:\begin{equation}\tilde{T}_v^{(l)} = \text{Concat}\left( \text{agg}_v^{(l)}, T_v^{(l)} \right)\label{eq:concat_operation}\end{equation}In this process, the node-enhanced token \(\text{agg}_v^{(l)}\), enriched with neighborhood information, is propagated to the subsequent text encoding layers to simultaneously perceive the topological associations of nodes in the graph when modeling semantics, avoiding text understanding from being detached from structural constraints.

\subsection{Text Encoding}

The text encoding module dynamically calculates the dependency weights between tokens in the sequence and with graph-enhanced tokens based on the Transformer, capturing contextual semantic associations. The specific formulas are:\begin{equation}\tilde{T}_v^{(l)}{}' = \text{LN}\left( T_v^{(l)} + \text{MHAasy}_l\left( \tilde{T}_v^{(l)} \right) \right)\label{eq:text_encoding_1}\end{equation}\begin{equation}T_v^{(l+1)} = \text{LN}\left( \tilde{T}_v^{(l)}{}' + \text{MLP}\left( \tilde{T}_v^{(l)}{}' \right) \right)\label{eq:text_encoding_2}\end{equation}where MLP is the multi-layer perceptron, LN is layer normalization, \(\text{MHAasy}\) is asymmetric multi-head attention, and the output sequence \(T_v^{(l+1)}\) serves as the input to the next layer. The last text encoding module outputs the [cls] token \(T_j^{(L)}[\text{cls}]\) as the final representation of the corresponding node, which is used in downstream tasks.

\subsection{Simple Aggregation Strategies}
As a key component of Odin's hierarchical alternating architecture, the choice of simple aggregation method for TS layers directly determines the transmission quality of structural information in non-key semantic stages — improper strategy design may lead to interruption, lag, or redundancy of structural information, thereby reducing the overall model performance. To this end, we design four differentiated simple aggregation strategies combined with the TG layer position configuration logic and semantic hierarchical requirements. Their performance and adaptability are systematically evaluated through extensive experiments.

Strategy 1 is VA (Vain, no processing). This strategy does not introduce any structural information in the TS layer, directly passing the text embedding to the next BERT layer, and only relies on the text encoding module itself to complete semantic iterative update. Its core feature is extremely low computational overhead, but it may cause text semantic modeling to deviate from graph topology constraints due to the lack of structural guidance.

Strategy 2 is ME (Mean, mean aggregation). In the TS layer, the mean function is directly used to aggregate the node and its neighbors. The specific formula is:
\begin{equation*}
	\begin{aligned}
		\text{agg}_v^{(l)} = \text{mean}\left( T_v^{(l)}[\text{cls}] \cup \left\{ T_j^{(l)}[\text{cls}] \mid j \in \mathcal{N}(v) \right\} \right)
	\end{aligned}
\end{equation*}

This strategy realizes mild injection of local structural information through parameter-free mean operation, which can not only avoid the over-smoothing risk caused by trainable weights in traditional GNN aggregation but also maintain the basic association between structure and semantics at low computational cost.

Strategy 3 is PE (Pre-order Enhanced Label Reuse). In the TS layer, the node-enhanced token \(\text{agg}_v^{(m)}\) generated by the previous TG layer is used as the node-enhanced token of the current layer. The specific formula is:

\begin{equation*}
	\begin{aligned}
		\text{agg}_v^{(l)} = \text{agg}_v^{(m)}
	\end{aligned}
\end{equation*}
where m denotes the layer number of the previous TG layer. Its core idea is to reuse the generated structural enhancement tokens without re-aggregating the neighborhood of the current layer, which can minimize the computational overhead of neighborhood sampling and aggregation, but may reduce the semantic adaptability with the current layer due to the lag of structural information (the enhanced token comes from the pre-order semantic layer).

Strategy 4 is PG (Pre-order GNN Layer Reuse). In the TS layer, the graph aggregation module parameters generated by the previous TG layer are used to undertake the aggregation task. The specific formula is:
\begin{equation*}
	\begin{aligned}
		\text{agg}_v^{(l)} = W_1^{(m)} \cdot \text{mean}\left( \left\{ T_j^{(l)}[\text{cls}] \mid j \in \mathcal{N}(v) \right\} \right) + W_2^{(m)} \cdot T_v^{(l)}[\text{cls}]
	\end{aligned}
\end{equation*}
where m denotes the layer number of the previous TG layer. This strategy ensures that the structural aggregation logic of the TS layer is consistent with that of the TG layer, improving the coherence of structural information transmission, while avoiding parameter redundancy caused by separately designing a graph aggregation module for the TS layer.

\subsection{Lightweight Odin}

When generating node embeddings, the standard Odin needs to generate embeddings for all current nodes and their sampled neighborhoods, leading to a linear increase in computational cost with the neighborhood scale and a significant efficiency bottleneck in large-scale graph inference scenarios. To mitigate this issue, we propose Lightweight Odin, which inserts graph aggregation modules into selected layers of a 6-layer Transformer as the primary aggregation stages, while employing the same secondary simple-aggregation strategy in all remaining layers. Compared with the standard Odin, the lightweight version has a slight performance loss but significantly improved inference speed, making it more suitable for resource-constrained application scenarios.

\subsection{Odin Forward Algorithm}
Algorithm 1 describes the embedding generation process when the entire graph $G(V, E)$ and the features of all nodes $v \in V$ are used as inputs. First, GraphSAGE minibatch is used to sample an $A$-hop subgraph centered on the nodes in the current batch. Then, the initial embeddings of the nodes are used as the current embeddings of the nodes, which are fed into the $L$-layer Odin and processed alternately by the aggregation layers and text encoding layers. In the aggregation layer, the [cls] token of the node is fed into the graph aggregation module (in the TG layer) or the simple aggregation module (in the TS layer) to generate the graph-enhanced token with graph structural information. Subsequently, the processed graph-enhanced token is distributed to its original node and concatenated with the original embedding to generate the graph-enhanced token-level embedding; the graph-enhanced token-level embedding is further processed by the subsequent text encoding layer to generate an embedding with higher-level semantics, which is then fed into the next Odin layer.
\begin{algorithm}[!htbp]
	\small
	\centering
	\caption{Embedding Generation Process of Odin}
	\label{alg:odin_embedding}
	\begin{algorithmic}[1]
		\Require
		Graph $G(V, E)$;
		Text features $x_v$ for all nodes $v \in V$;
		Current batch of nodes $B$;
		Layers $K$;
		Hop $A$;
		Distribution Setting of TG Layers $S$;
		Asymmetric Multi-Head Attention $\text{MHAasy}_l$;
		Multi-Layer Perceptron $\text{MLP}_l$;
		Weight matrices $W_1^l$, $W_2^l$;
		Non-linearity $\sigma$;
		$\text{SimpleAggregate}(\cdot)$;

		\Ensure
		Output: The output of the last hidden layer $T_v^L[\text{cls}]$ for all node $v \in B$;
		
		\State $m \leftarrow 0$
		\State $B_A \leftarrow B$
		
		\For{$a = A$ \textbf{downto} $1$} \text{// GraphSAGE's Minibatch Sampling}
		\State $B_{a-1} \leftarrow \text{NeighborhoodSampling}(B_{a}) \cup B_{a} $
		\EndFor
		
		\For{ $v \in  B_0$}
		\State $T_v^0 \leftarrow \text{InitialTokenGeneration}(x_v)  \text{// Initial Token Generation}$;
		\State $\tilde{T}_v^0{}' = \text{LN}\left(T_v^0 + \text{MHAasy}_l\left(T_v^0\right)\right)   \text{  //Eq.(3)}$
		\State $T_v^1 = \text{LN}\left(\tilde{T}_v^0{}' + \text{MLP}_l\left(\tilde{T}_v^0{}'\right)\right)\text{  //Eq.(4)}$
		\EndFor
		
		\For{$l = 1$ \textbf{to} $K-1$}
		\For{$v \in  B_m$}
		\If{$l \in S$}
		\State $\text{agg}_v^l = W_1^l \cdot \text{mean}\left( \left\{ T_j^l[\text{cls}]  // \mid j \in \mathcal{N}(v) \right\} \right) + W_2^l \cdot T_v^l \text{  //Eq.(1a)}$
		
		\Else
		\State $\text{agg}_v^l = \text{SimpleAggregate}\left( T_v^l[\text{cls}], \left\{ T_j^l[\text{cls}] \mid j \in \mathcal{N}(v) \right\} \right) \text{  //Eq.(1b)}$
		\EndIf
		\State $\tilde{T}_v^l = \text{Concate}\left( \text{agg}_v^l, T_v^l \right)$
		\State $\tilde{T}_v^l{}' = \text{LN}\left( \tilde{T}_v^l + \text{MHAasy}_l\left( \tilde{T}_v^l \right) \right) \text{  //Eq.(3)}$
		\State $T_v^{l+1} = \text{LN}\left( \tilde{T}_v^l{}' + \text{MLP}_l\left( \tilde{T}_v^l{}' \right) \right) \text{  //Eq.(4)}$
		\EndFor
		\If{$l \in S$}
		\State $m \leftarrow m + 1$
		\EndIf
		\EndFor
	\end{algorithmic}
	
\end{algorithm}

\subsection{Odin Training Process}

Considering the strong correlation between text semantics and graph structure in text-attributed graphs, we refer to two pre-training strategies proposed by PATTON [24]: Masked Node Prediction (MNP) and Network Context-Aware Masked Language Modeling (NMLM). Through these two self-supervised objectives, we jointly capture token-level text semantics and document-level structural associations. The formula for Masked Node Prediction (MNP) is:\begin{equation}L_1 = -\sum_{v_j \in M_v} \sum_{v_k \in \mathcal{N}(v_j)} \log \frac{\exp\left(T_{v_j}[\text{cls}]^T T_{v_k}[\text{cls}]\right)}{\exp\left(T_{v_j}[\text{cls}]^T T_{v_k}[\text{cls}]\right) + \exp\left(T_{v_j}[\text{cls}]^T T_{v'_u}[\text{cls}]\right)}\end{equation}where \(M_v\) is the set of masked nodes, \(\mathcal{N}(v_j)\) denotes the set of neighbors of node \(v_j\), \(T_v[\text{cls}]\) represents the embedding of node \textit{v} in the output of the model’s last hidden layer, and \(v'_u\) denotes negative sample nodes in the current batch that are not connected to \(v_j\). 

The formula for Network Context-Aware Masked Language Modeling (NMLM) is:\begin{equation}L_2 = -\sum_{i \in M_t} \log \text{softmax}\left(q_{w_i}^T T_i\right)\end{equation}where \(T_i\) is the hidden state of the masked token (belonging to node \textit{i}) in the last layer of the model, \(M_t\) is the set of all tokens processed by masking, \(w_i\) is the masked token, and \(q_{w_i}\) is the MLM prediction head for \(w_i\). 

The total pre-training loss is defined by weighting and combining the NMLM and MNP objectives:\begin{equation}L = L_1 + L_2\end{equation}The fine-tuning stage for downstream tasks in this paper is consistent with PATTON [24], and a differentiated encoding scheme is designed for two input scenarios (intra-network text and extra-network text). For details, refer to \citep{ref24}.

\subsection{Theoretical Analysis}
\parindent=0pt
This section establishes the key theoretical properties that distinguish Odin from conventional GNN- or Transformer-based models. While deep GNNs suffer from oversmoothing and pure Transformers lack relational inductive bias, Odin’s hierarchical fusion mechanism integrates structural and semantic information in a fundamentally different way. To formalize these advantages, we present two core results: Theorem 1 shows that Odin inherently avoids oversmoothing, and Theorem 2 proves that Odin’s expressive power strictly subsumes that of both GNNs and Transformers.

\textbf{Theorem 1 (Oversmoothing Avoidance). } The hierarchical fusion architecture of Odin inherently avoids the oversmoothing phenomenon that affects deep GNNs.

\textbf{Proof. }The smoothness of an M-layer GNN (e.g., GraphSAGE) converges as M increases, eventually collapsing into identical vectors (i.e., oversmoothing). In contrast, Odin performs aggregation not over GNN-computed hidden states but over the [CLS] tokens produced by the Transformer at each layer. Crucially, [CLS] representations do not exhibit smoothing behavior. The residual connections and multi-head attention mechanisms that govern their evolution do not constitute contraction or averaging operators; attention selectively redistributes importance across tokens rather than homogenizing them. Because Odin aggregates these non-smoothing [CLS] representations at every structural injection step, the aggregation pipeline never introduces the iterative averaging dynamics responsible for GNN oversmoothing. Therefore, no sequence of Odin layers produces convergence toward constant vectors. $\square$

\parindent=15pt
This theorem establishes that Odin’s hierarchical fusion structure fundamentally circumvents GNN oversmoothing. Unlike deep GNNs—whose representations collapse into homogeneous embeddings as depth increases—Odin leverages Transformer-derived [CLS] tokens, whose dynamics remain expressive due to residual pathways and attention’s selective weighting. Consequently, Odin preserves discriminative and non-degenerate node features even at substantial depth, providing theoretical assurance for its robustness in deep graph–text fusion.

\parindent=0pt
\textbf{Theorem 2 (Expressive Power of Odin). } 
\parindent=15pt
The expressive power of Odin strictly contains that of both Transformers and GNNs. 
 
The proof of Theorem 2 is detailed in Appendix A.

Together, these theorems provide strong theoretical justification for Odin’s design. By avoiding the collapse behavior of deep GNNs and exceeding the expressive limits of existing architectures, Odin offers a principled foundation for deep structure–semantic integration.

\section{Experimentation}
To systematically verify the effectiveness of the proposed Odin framework in text-attributed graph representation learning, this section conducts analysis through multi-dataset and multi-task experiments based on the following 4 verifiable sub-questions:
\begin{itemize}\item \textbf{Q1}: How do the number and insertion positions of TG layers affect model performance in the Odin architecture?\item \textbf{Q2}: Which simple aggregation strategy can maximize the transmission efficiency of structural information and thereby optimally improve Odin’s performance?\item \textbf{Q3}: Can Odin with a 6-layer Transformer block as the text encoder also achieve excellent performance? What about its efficiency?\item \textbf{Q4}: In the four downstream tasks (node classification, link prediction, retrieval, and re-ranking), how does Odin perform compared with mainstream pre-trained language models (PLMs), graph neural networks (GNNs), and traditional GNN-LM fusion models?\end{itemize}

\subsection{Data Preparation and Processing}
We use five widely used text-attributed graph datasets: Cora \citep{ref45}, CiteSeer \citep{ref19}, ArXiv2023 \citep{ref20}, OGBN-Products \citep{ref21}, and OGBN-ArXiv \citep{ref21}. Due to the large scale of OGBN-ArXiv, this work uses the Breadth-First Search (BFS) algorithm to extract three subgraphs with 3,000, 6,000, and 10,000 nodes from OGBN-ArXiv as experimental datasets, denoted as OGBN-ArXiv-3000, OGBN-ArXiv-6000, and OGBN-ArXiv-10000. The statistical information of the datasets is shown in Table 2.

\begin{table}[!htbp]
	\centering
	\caption{Statistical Information of Datasets}
	\label{tab:dataset_stats}
	\begin{tabular}{p{2.2cm} c c c c}
		\toprule
		Dataset & Nodes & Edges & Fine-Grained Classes & Merged Coarse-Grained Classes \\
		\midrule
		OGBN-ArXiv-3000 & 3,000 & 8,480 & 40 & 9 \\
		OGBN-ArXiv-6000 & 6,000 & 20,238 & 40 & 9 \\
		OGBN-ArXiv-10000 & 10,000 & 74,010 & 40 & 9 \\
		Cora & 2,708 & 5,429 & 7 & 7 \\
		CiteSeer & 3,186 & 4,277 & 6 & 6 \\
		ArXiv2023 & 46,198 & 78,543 & 40 & 9 \\
		OGBN-Products & 54,025 & 74,420 & 47 & 11 \\
		\bottomrule
	\end{tabular}
\end{table}

Since the Cora and CiteSeer datasets lack fine-grained classes that can be used as retrieval labels, experiments on the downstream tasks of retrieval and re-ranking are not conducted for them. For datasets supporting these two tasks (ArXiv2023, OGBN-Products, and OGBN-ArXiv subgraphs), the node labels of ArXiv2023 and OGBN-ArXiv are abbreviations; we expand them into corresponding full words (e.g., expanding the original dataset class name ``cs.AI" to ``Artificial Intelligence") to adapt to language model processing. In the classification task, similar fine-grained classes are merged into coarse-grained classes through topic relevance clustering; see Appendix B for the detailed merging table.

\subsection{Pre-training Settings}

The model is trained on one NVIDIA L20 GPU with a pre-training batch size of 32 and 10 training epochs. During the model pre-training stage, following the settings of PATTON \citep{ref24}, the dataset is split into training/validation/test sets at an 80\%/10\%/10\% ratio. The learning rate of the text encoder is set to 1e-5, and the learning rate of the graph aggregation module is set to 1e-3. The Network Context-Aware Masked Language Modeling (NMLM) pre-training adopts a standard 15\% [MASK] masking ratio. For the non-lightweight Odin, we use BERT-base \citep{ref17} (on the OGBN-Products dataset) and SciBERT \citep{ref23} (on the ArXiv2023, OGBN-Products, OGBN-ArXiv subgraphs, Cora, and CiteSeer datasets) as the initial checkpoints for the text encoding layer. For the lightweight Odin, we use DistilBERT \citep{ref18} as the initial checkpoint. During pre-training, one-hop neighbors need to be sampled for each TG layer; the number of sampled neighbors is an adjustable parameter, and the minibatch sampling in Odin’s pre-training uses 5 neighbors per hop.

\subsection{Baselines}

We mainly compare Odin with three categories of baselines. Odin and all baseline models are trained and tested with three different random number seeds in the four downstream tasks of the formal comparative experiments, with the mean and standard deviation reported.
\begin{enumerate}\item \textbf{LM-GNN Fusion Models}: Including \textbf{Patton} \citep{ref24}, cascaded \textbf{BERT+GraphSAGE} \citep{ref16,ref17}, \textbf{Patton-ckpt} \citep{ref24}, and \textbf{LLaGA} {ref2}. Patton is a GNN-LM fusion model based on the GraphFormers architecture, representing the LLM-as-Aligner category. PLM+SAGE is a classic cascaded GNN-LM fusion model, representing the text-structure serial modeling paradigm and belonging to the LLM-as-Enhancer category. Patton-ckpt {ref24} is a pre-trained Patton model provided by the Patton authors. LLaGA is an LM-GNN fusion model for large-scale text-attributed graphs, which achieves cross-modal feature alignment by introducing a lightweight projector and represents the LLM-as-Predictor category. Among them, Patton and PLM+SAGE undergo the same pre-training and downstream fine-tuning as our model; Patton-ckpt is a pre-trained checkpoint provided by the Patton authors, which we load directly for fine-tuning without additional pre-training. LLaGA uses Vicuna for initialization and only conducts training and testing for classification tasks.
	\item \textbf{Off-the-Shelf Pre-trained LM Models}: Including \textbf{SciBERT} \citep{ref23}, \textbf{LinkBERT} \citep{ref26}, \textbf{DistilBERT} \citep{ref18}, and \textbf{ModernBERT} \citep{ref25}. SciBERT is a domain-adapted pre-trained language model for scientific text, focusing on improving text semantic understanding in academic scenarios. LinkBERT is a pre-trained language model that incorporates document link information, aiming to learn cross-document semantic associations. DistilBERT is a knowledge-distilled lightweight version of BERT, focusing on balancing model efficiency and performance. ModernBERT is the latest general pre-trained language model, representing the state-of-the-art in pure text semantic modeling. For these models, we directly load their checkpoints for fine-tuning without additional pre-training.
	\item \textbf{GNN Models}: \textbf{GraphSAGE} \citep{ref16} is a typical inductive graph convolutional neural network algorithm and a classic baseline for graph structural modeling. \textbf{GraphSAGE++} \citep{ref27} is a performance-optimized version of GraphSAGE, aiming to improve the model’s ability to distinguish neighborhood information. For these models, we use SciBERT vectors to initialize and generate initial embeddings for each node. In link prediction experiments, the number of negative samples is set to 3 during training and 7 during testing. In classification experiments, the models are trained and tested on the same classification dataset. Since pure GNNs cannot generate fine-grained semantic representations, they only participate in node classification and link prediction tasks, not in retrieval and re-ranking tasks.
\end{enumerate}

\subsection{Model Architecture Selection}

Odin only designates key layers as TG layers; the number and positions of TG layers in the L-layer Odin directly affect model performance. Inserting TG layers too early may lead to the dilution of graph structural information by subsequent text encoding modules, while inserting them too late fails to guide semantic modeling based on the hierarchical characteristics of text-attributed graphs. Early text encoding layers typically focus on learning local grammatical rules and basic lexical associations; middle layers gradually achieve phrase-level semantic integration; late layers complete global semantic association and entity-level feature extraction. For Odin with 12 layers, this paper designs six differentiated TG layer position configurations (Table 3) to experimentally verify the guidance effect of structural information on full-stage semantic modeling under different TG layer configurations. To determine the optimal architectural parameters of Odin, we first test the model performance on the Cora dataset: all models with different TG layer settings undergo 10 pre-training epochs on Cora, with 32-shot fine-tuning for link prediction and 8-shot fine-tuning for classification tasks. The experimental results are shown in Table 3.
\begin{table}[h]
	\centering
	
	\caption{Experimental Results of Models with Different TG Layer Settings on the Cora Dataset}
	\label{tab:mean-friedman}
	\begin{tabular}{lcc}
		\toprule
		TG Layer Positions & Link Prediction (PREC) & Classification (ACC) \\
		\midrule
		1, 2, 3          & 0.735±0.004            & 0.657±0.021           \\
		9, 10, 11        & 0.736±0.008            & 0.677±0.004           \\
		1, 6, 11         & \underline{0.783±0.031}            & \underline{0.724±0.009}         \\
		3, 6, 9          & 0.739±0.007            & 0.668±0.009           \\
		4, 8             & 0.749±0.010            & 0.663±0.005           \\
		1, 4, 8, 11      & \textbf{0.796±0.031}            & \textbf{0.724±0.003}           \\
		\bottomrule
	\end{tabular}
\end{table}
For the lightweight Odin with 6 layers (L=6), we set four different TG layer position configurations (Table 4) to verify the guidance effect of structural information on full-stage semantic modeling. We then test the performance of the lightweight models on the same dataset with the same experimental settings; the results are shown in Table 4.

\begin{table}[h]
	\centering
	
	\caption{Experimental Results of Models with Different TG Layer Settings on the Cora Dataset}
	\label{tab:mean-friedman}
	\begin{tabular}{lcc}
		\toprule
		TG Layer Positions & Link Prediction (PREC) & Classification (ACC) \\
		\midrule
		5          & 0.664±0.001             & 0.657±0.005          \\
		2          & 0.661±0.001             & \underline{0.676±0.008}           \\
		1,3,5      & \underline{0.683±0.002}             & 0.615±0.010         \\
		2,4        & \textbf{0.702±0.032}             & \textbf{0.668±0.005}           \\
		\bottomrule
	\end{tabular}
\end{table}

After selecting the optimal TG layer settings, we test four different simple aggregation strategies on Odin and lightweight Odin using the Cora dataset; the results are shown in Table 5.
\begin{table}[htbp]
	\centering
	\caption{Experimental Results of Models with Different simple aggregation Layer Treatments on the Cora Dataset}
	\label{tab:non_agg_layer_results}
	\begin{tabular}{lcc}
		\toprule
		non-main aggregation layers Treatments & Link Prediction (PREC) & Classification (ACC) \\
		\midrule
		Odin TG-1,6,11 VA        & 0.743±0.005      & 0.663±0.014 \\
		Odin TG-1,6,11 PG        & \textbf{0.783±0.031}      & \textbf{0.724±0.009} \\
		Odin TG-1,6,11 ME        & 0.744±0.007      & \underline{0.664±0.022}   \\
		Odin TG-1,6,11 PE        & 0.742±0.005      & 0.648±0.013            \\
		Odin TG-ALL 3jump        & \underline{0.747±0.001}      & 0.657±0.011            \\
		Odin TG-ALL 2jump        & 0.747±0.001      & 0.653±0.009            \\
		Odin TS-ALL 3jump        & 0.423±0.003      & 0.471±0.009            \\
		\midrule
		OdinLight TG-2,4 VA      & 0.590±0.027      & 0.555±0.002   \\
		OdinLight TG-2,4 PG      & \textbf{0.702±0.032}      & \textbf{0.684±0.005} \\
		OdinLight TG-2,4 ME      & 0.637±0.008      & 0.585±0.022   \\
		OdinLight TG-2,4 PE      & 0.664±0.020      & 0.596±0.008             \\
		OdinLight TG-ALL 3jump   & \underline{0.676±0.004}      & \underline{0.620±0.002} \\
		\midrule
		OdinLight TG-2 VA        & \underline{0.596±0.005}      & \underline{0.568±0.003}   \\
		OdinLight TG-2 PG        & \textbf{0.661±0.014}      & \textbf{0.676±0.008}   \\
		OdinLight TG-2 ME        & 0.583±0.015      & 0.544±0.005   \\
		OdinLight TG-2 PE        & 0.538±0.028      & 0.478±0.006   \\
		\bottomrule
	\end{tabular}
\end{table}

Among them, ``TG-ALL 3jump" refers to Odin where all layers are TG layers but the sampled subgraphs only include 3-hop neighbors; ``TG-ALL 2jump" refers to Odin where all layers are TG layers but the sampled subgraphs only include 2-hop neighbors; ``TS-ALL 3jump" refers to Odin where all layers are TS layers but the sampled subgraphs only include 3-hop neighbors. These three variants are used to verify the role of simple aggregation strategies in improving model performance.
\paragraph*{Regarding Q1: How do the number and insertion positions of TG layers affect model performance?}Based on the above experimental results on the Cora dataset, the number and insertion positions of TG layers have a significant impact on Odin's performance, and the selection needs to be adapted to the model architecture. The results show that for Odin with L=12, the model achieves optimal performance when TG layers are at [1,4,8,11] and [1,6,11], which is 5\%-6\% higher than the worst configuration. The reason is that uniform distribution ensures structural information plays a role in the full ``syntax-semantics" stage of the text encoding module, avoiding information imbalance caused by insertion in a single stage. Although [1,4,8,11] achieves the best performance, it requires sampling 4-hop neighbors, which brings huge computational costs and risks of out-of-memory during training on large datasets. Therefore, we select the sub-optimal [1,6,11] as the setting for comparative experiments. For lightweight Odin with L=6, the model performs well when TG layers are at [2,4]. We use [2,4] and [2] as the settings for lightweight Odin in comparative experiments.
\paragraph*{Regarding Q2: Which simple aggregation strategy can maximize the transmission efficiency of structural information and thereby optimally improve Odin’s performance?}From the experimental results, in non-lightweight Odin, PG perform well, and their performance is better than TG-ALL 3jump. The poor performance of TG-ALL is due to excessive graph aggregation modules in full TG layer configuration, leading to excessive parameters and over-injection of structural information, which causes overfitting and performance degradation. We select PG, ME, and VA as the official experimental models. For the lightweight Odin model, the PG method performs the best and is therefore chosen as its official experimental model.

\subsection{Link Prediction}
In this section, we perform 32-shot link prediction fine-tuning on nodes in the network. Unlike the pre-training stage, Odin training in this stage does not include the NMLM task. We use the hidden state of the [CLS] token in the last layer as the node representation and conduct in-batch evaluations. The experimental results are shown in Table 6. It can be observed that the non-lightweight Odin outperforms baseline models in most cases, and the lightweight Odin also achieves good performance. This indicates that the proposed pre-training strategy helps language models extract knowledge from text-rich pre-trained networks and apply it to edge type prediction tasks.

\begin{table}[!htbp]
	\centering
	\footnotesize
	\renewcommand{\arraystretch}{1.3}
	\setlength{\tabcolsep}{3pt}
	\caption{PRC Values of Link Prediction Task on Various Datasets}
	\label{tab:link_pred_prc_all_datasets}
	\begin{tabular*}{\linewidth}{@{\extracolsep{\fill}} cccccccc}
		\toprule
		 Model & O3k & O6k & O10k & Cora & CiteS & Arx2023 & Products\\
		\midrule
		 
		 Odin VA      & \underline{0.564±0.02} & \underline{0.638±0.03} & \textbf{0.573±0.03}    & 0.743±0.01             & 0.806±0.01             & \underline{0.774±0.02} & \textbf{0.948±0.00} \\
		 Odin PG      & 0.549±0.05             & 0.571±0.01             & \underline{0.564±0.01} & \textbf{0.783±0.03} & 0.807±0.01             & \textbf{0.780±0.02} & \underline{0.946±0.00} \\
		 Odin ME      & \textbf{0.609±0.05}    & \textbf{0.677±0.02}    & 0.563±0.02             & 0.744±0.007    & \underline{0.810±0.02} & 0.768±0.00 & 0.942±0.00 \\
		 Odin-L 2,4PG & 0.552±0.03             & 0.556±0.05             & 0.540±0.03             & 0.702±0.03             & 0.748±0.03             & 0.732±0.03 & 0.708±0.00 \\
		 Odin-L 2PG   & 0.536±0.02             & 0.508±0.06             & 0.529±0.02             & 0.661±0.01             & 0.738±0.04             & 0.717±0.04 & 0.607±0.00 \\
		 PLM+SAGE     & 0.501±0.01             & 0.552±0.02             & 0.524±0.01             & 0.724±0.03             & 0.708±0.02             & 0.679±0.00 & 0.537±0.02 \\
		 PATTON       & 0.515±0.01             & 0.608±0.01             & 0.545±0.01             & \underline{0.758±0.02}             & \textbf{0.814±0.02}    & 0.767±0.01 & 0.941±0.00 \\
		 PATTON-PT    & 0.443±0.00             & 0.469±0.01             & 0.436±0.00             & 0.608±0.02             & 0.632±0.01             & 0.558±0.01 & 0.709±0.00 \\
		\midrule
		
		 ModernBERT & 0.220±0.04 & 0.209±0.01 & 0.175±0.03 & 0.258±0.03 & 0.360±0.03 & 0.390±0.01 & 0.557±0.01 \\
		 DistillBERT & 0.127±0.00 & 0.125±0.00 & 0.125±0.00 & 0.124±0.00 & 0.129±0.00 & 0.126±0.00 & 0.133±0.00 \\
		 LinkBERT  & 0.143±0.01 & 0.136±0.01 & 0.132±0.00 & 0.140±0.01 & 0.157±0.01 & 0.139±0.00 & 0.198±0.00 \\
		 SciBERT   & 0.313±0.02 & 0.310±0.02 & 0.259±0.00 & 0.514±0.03 & 0.451±0.00 & 0.520±0.00 & 0.539±0.02 \\
		\midrule
		
		 GraphSAGE & 0.528±0.07 & 0.348±0.03 & 0.152±0.02 & 0.162±0.03 & 0.221±0.01 & 0.511±0.05 & 0.155±0.01 \\
		 GraphSAGE++ & 0.559±0.03 & 0.507±0.08 & 0.543±0.05 & 0.213±0.03 & 0.286±0.07 & 0.494±0.05 & 0.201±0.01 \\
		\bottomrule
	\end{tabular*}
	\vspace{0.3em}
	\parbox{\linewidth}{\tiny Note: O3k=Obgn3000, O6k=Obgn6000, O10k=Obgn10000, CiteS=CiteSeer, Odin-L=Odin Light, PT=Pretrain, Arx2023=Arxiv2023. Values rounded to 3 decimal places.}
\end{table}

\subsection{Node Classification}
In this section, we conduct 8-shot coarse-grained classification experiments on nodes in the network. We use the hidden state of the [CLS] token in the last layer output by the language model as the node representation and input it into a linear layer classifier to obtain prediction results. The experimental results are shown in Table 7. Both the language model and the classifier are fine-tuned. As shown in Table 7, the non-lightweight Odin outperforms baseline models in most cases; the lightweight Odin shows average performance but still achieves a significant improvement compared to its initial checkpoint (DistilBERT).
\begin{table}[!htbp]
	\centering
	\footnotesize
	\setlength{\tabcolsep}{3pt}
	\caption{ACC Values of Node Classification Task on Various Datasets}
	\label{tab:link_pred_prc_all_datasets}
	\begin{tabular*}{\linewidth}{@{\extracolsep{\fill}} cccccccc}
		\toprule
		 Model & O3k & O6k & O10k & Cora & CiteS & Arx2023 & Products\\
		\midrule
		 Odin VA      & \underline{0.576±0.01} & \textbf{0.578±0.01}    & 0.541±0.01             & 0.663±0.01             & 0.601±0.01 & 0.566±0.00 & \underline{0.700±0.00}\\
		 Odin PG      & 0.566±0.02             & \underline{0.567±0.01} & 0.511±0.01             & 0.664±0.022 & 0.607±0.01 & \underline{0.581±0.00} & \textbf{0.700±0.01} \\
		 Odin ME      & \textbf{0.582±0.01}    & 0.549±0.02             & \underline{0.542±0.02} & \textbf{0.729±0.00}    & 0.596±0.01 & \textbf{0.582±0.01} & 0.695±0.00 \\
		 Odin-L 2,4PG & 0.371±0.01             & 0.418±0.02             & 0.506±0.03             & 0.684±0.03             & 0.567±0.01 & 0.416±0.01 & 0.521±0.03 \\
		 Odin-L 2PG   & 0.351±0.01             & 0.365±0.02             & 0.487±0.02             & 0.676±0.00             & 0.594±0.01 & 0.421±0.01 & 0.457±0.01 \\
		 PLM+SAGE     & 0.339±0.01             & 0.389±0.02             & \textbf{0.558±0.01}    & \underline{0.691±0.01}             & 0.630±0.01 & 0.452±0.03 & 0.490±0.01 \\
		 PATTON       & 0.518±0.01             & 0.455±0.01       l      & 0.520±0.06             & 0.678±0.02             & \underline{0.642±0.01} & 0.553±0.02 & 0.672±0.01 \\
		 PATTON-PT    & 0.391±0.01             & 0.452±0.01             & 0.487±0.06             & 0.625±0.01             & \textbf{0.661±0.00} & 0.489±0.00 & 0.597±0.02 \\
		\midrule
		
		 ModernBERT & 0.117±0.01 & 0.186±0.02 & 0.229±0.02 & 0.229±0.02 & 0.224±0.02 & 0.215±0.01 & 0.382±0.01 \\
		 DistillBERT & 0.061±0.03 & 0.167±0.23 & 0.050±0.02 & 0.084±0.02 & 0.199±0.02 & 0.041±0.00 & 0.093±0.01 \\
		 LinkBERT  & 0.145±0.01 & 0.197±0.02 & 0.097±0.02 & 0.155±0.02 & 0.182±0.01 & 0.120±0.04 & 0.203±0.01 \\
		 SciBERT   & 0.264±0.03 & 0.190±0.02 & 0.361±0.06 & 0.450±0.06 & 0.489±0.02 & 0.460±0.01 & 0.409±0.02 \\
		\midrule
		
		 GraphSAGE & 0.301±0.04 & 0.368±0.03 & 0.311±0.01 & 0.571±0.03 & 0.533±0.02 & 0.340±0.02 & 0.458±0.01 \\
		 GraphSAGE++ & 0.312±0.02 & 0.372±0.04 & 0.327±0.04 & 0.524±0.02 & 0.500±0.00 & 0.352±0.00 & 0.463±0.002 \\
		\bottomrule
	\end{tabular*}
	\vspace{0.3em}
	\parbox{\linewidth}{\tiny Note: O3k=Obgn3000, O6k=Obgn6000, O10k=Obgn10000, CiteS=CiteSeer, Odin-L=Odin Light, PT=Pretrain,Arx2023=Arxiv2023; Values rounded to 2 decimal places.}
\end{table}

\subsection{Retrieval}
The retrieval task corresponds to 16-shot fine-grained category retrieval: given a node, we need to retrieve its category name from a large label space. We adopt the widely used DPR (Dense Passage Retrieval, (Karpukhin et al., 2020)) pipeline to fine-tune all models. Specifically, the last-layer hidden state of the [CLS] token is used as the dense representation of nodes and category names. We use the BM25 algorithm to assign negative samples to nodes. The experimental results are shown in Table 8. It can be seen from the results that non-lightweight Odin outperforms all baseline methods, while lightweight Odin performs moderately. This may be due to the lack of scientific literature vocabulary in its initial checkpoint, leading to poor generalization ability. However, it still achieves a significant performance improvement compared to DistillBERT, its initial checkpoint.

\begin{table}[!htbp]
	\centering
	\renewcommand{\arraystretch}{1.2} 
	\footnotesize
	\setlength{\tabcolsep}{4pt} 
	\caption{Recall@10 Values of Retrieval Task on Various Datasets}
	\label{tab:retrieval_recall10}
	\begin{tabular*}{\linewidth}{@{\extracolsep{\fill}} lccccc}
		\toprule
		 Model & Obgn3000 & Obgn6000 & Obgn10000 & Arxiv2023 & Obgn-produce \\
		\midrule
		
		 Odin VA             & 0.927±0.01             & \underline{0.891±0.02} & 0.861±0.05             & \textbf{0.932±0.00}    & 0.861±0.01 \\
		 Odin PG             & \underline{0.928±0.02} & 0.815±0.02             & \underline{0.916±0.02} & \underline{0.932±0.01} & \underline{0.863±0.01} \\
		 Odin ME             & \textbf{0.931±0.01}    & \textbf{0.912±0.02}    & \textbf{0.924±0.01}    & 0.913±0.01             & \textbf{0.869±0.01} \\
		 Odin-L 2,4PG & 0.271±0.04             & 0.354±0.08             & 0.234±0.24             & 0.214±0.16             & 0.483±0.01 \\
		 Odin-L 2PG   & 0.264±0.02             & 0.286±0.22             & 0.232±0.23             & 0.124±0.01             & 0.471±0.00 \\
		 PLM+sage                 & 0.816±0.02             & 0.737±0.01             & 0.894±0.00             & 0.868±0.00             & 0.616±0.03 \\
		 PATTON                   & 0.859±0.03             & 0.807±0.04             & 0.859±0.03             & 0.906±0.01             & 0.830±0.01 \\
		 PATTON-Pretrain          & 0.796±0.06             & 0.752±0.06             & 0.880±0.04             & 0.876±0.02             & 0.640±0.03 \\
		\midrule
		
		 ModernBERT            & 0.520±0.05 & 0.218±0.14 & 0.329±0.21 & 0.347±0.04 & 0.742±0.06 \\
		 DistillBERT           & 0.257±0.07 & 0.210±0.00 & 0.490±0.22 & 0.304±0.00 & 0.496±0.03 \\
		 LinkBERT              & 0.217±0.02 & 0.220±0.07 & 0.241±0.07 & 0.197±0.10 & 0.449±0.08 \\
		 SciBERT               & 0.503±0.04 & 0.763±0.05 & 0.332±0.01 & 0.289±0.08 & 0.616±0.02 \\
		\bottomrule
	\end{tabular*}
\end{table}

\subsection{Reranking}
The reranking task corresponds to 32-shot fine-grained category reranking. We first use BM25 (Stephen Robertson et al., 2008) and exact matching as retrievers to obtain a list of candidate category names for each node; then, the model is required to rerank all categories in the list based on the similarity between the candidate categories and the given node text. The encoding method for nodes and category names is consistent with the retrieval task.

Different from the retrieval task, the reranking task tests the language model's ability to distinguish candidate categories at the fine-grained level. The experimental results are shown in Table 9. It can be found from the results that non-lightweight Odin outperforms baseline models in most cases, while lightweight Odin performs moderately.

\begin{table}[!htbp]
	\centering
	\renewcommand{\arraystretch}{1.2}
	\footnotesize
	\setlength{\tabcolsep}{4pt}
	\caption{PRC Values of Reranking Task on Various Datasets}
	\label{tab:reranking_prc}
	\begin{tabular*}{\linewidth}{@{\extracolsep{\fill}} lccccc}
		\toprule
		 Model & Obgn3000 & Obgn6000 & Obgn10000 & Arxiv2023 & Obgn-produce \\
		\midrule
		
		 Odin VA     & 0.751±0.03             & \textbf{0.686±0.03}    & 0.701±0.02 & 0.656±0.02             & 0.612±0.00 \\
		 Odin PG     & \underline{0.752±0.02} & 0.669±0.04             & 0.717±0.02 & \underline{0.657±0.02} & \textbf{0.635±0.03} \\
		 Odin ME     & \textbf{0.757±0.01}    & \underline{0.680±0.04} & 0.709±0.02 & \textbf{0.664±0.01}    & \underline{0.618±0.06} \\
		 Odin-L 2,4PG  & 0.371±0.07             & 0.467±0.12             & 0.764±0.06 & 0.523±0.03             & 0.594±0.00 \\
		 Odin-L 2PG    & 0.420±0.09             & 0.486±0.13             & 0.728±0.07 & 0.452±0.12             & 0.592±0.00 \\
		 PLM+sage              & 0.606±0.00             & 0.601±0.01             & 0.718±0.01 & 0.551±0.02             & 0.496±0.02 \\
		 PATTON                & 0.736±0.02             & 0.656±0.05             & 0.701±0.01 & 0.632±0.02             & 0.607±0.02 \\
		 PATTON-Pretrain       & 0.556±0.19             & 0.450±0.16             & 0.744±0.02 & 0.582±0.01             & 0.458±0.06 \\
		\midrule
		
		 ModernBERT            & 0.344±0.05 & 0.355±0.07 & 0.421±0.08             & 0.421±0.08 & 0.587±0.05 \\
		 DistillBERT           & 0.081±0.10 & 0.402±0.00 & \textbf{0.802±0.00}    & 0.104±0.00 & 0.592±0.00 \\
		 LinkBERT              & 0.178±0.13 & 0.262±0.18 & \underline{0.773±0.04} & 0.242±0.11 & 0.172±0.06 \\
		 SciBERT               & 0.261±0.04 & 0.269±0.05 & 0.594±0.15             & 0.562±0.03 & 0.477±0.05 \\
		\bottomrule
	\end{tabular*}
\end{table}

\subsection{Pretraining Efficiency Comparison}
To systematically evaluate the training efficiency of each model, we counted the pretraining time of different models on the ARXIV2023 dataset (46,198 nodes, 78,543 edges). The results are shown in Table 10.

\begin{table}[!htbp]
	\centering
	\renewcommand{\arraystretch}{1.2}
	\small
	\setlength{\tabcolsep}{6pt}
	\caption{Pretraining Time on ARXIV2023 Dataset}
	\label{tab:pretraining_time}
	\begin{tabular*}{\linewidth}{@{\extracolsep{\fill}} lc}
		\toprule
		Model & Pretraining Time (ARXIV2023) \\
		\midrule
		Odin-1,6,11 VA       & 1h 12m 28s \\
		Odin-1,6,11 PG       & 1h 13m 26s \\
		Odin-1,6,11 ME       & 1h 14m 27s \\
		Patton               & 1h 41m 50s \\
		Odin-ALL 3jump       & 3h 2m 23s OOM \\
		OdinLight-2,4        & 40min 18s \\
		OdinLight-2          & 24m 51s \\
		PLM+sage             & 1h 38m 11s \\
		\bottomrule
	\end{tabular*}
	\vspace{0.3em}
	\parbox{\linewidth}{\tiny Note: Odin-ALL 3jump (all layers set as TG layers but only 3-hop node sampling) encountered out-of-memory (OOM) during training due to excessive neighborhood sampling volume, making it impossible to complete full training. We estimated the time required for its full pretraining using the Transformer library as a reference for this model's time consumption.}
\end{table}

It is particularly noted that Odin-ALL 3jump (all layers set as TG layers but only 3-hop node sampling) encountered out-of-memory (OOM) during training due to excessive neighborhood sampling volume, making it impossible to complete full training. We estimated the time required for its full pretraining using the Transformer library as a reference for this model's time consumption. From the results, it can be seen that the pretraining time of Odin is lower than that of Patton, PLM+sage, and Odin-ALL 3jump, indicating that the use of a simple aggregation strategy can not only improve Odin's performance but also reduce its training cost. On this basis, the pretraining time of lightweight Odin is further reduced. Among them, lightweight Odin with fewer TG layers has lower pretraining cost, because lightweight Odin with fewer TG layers requires sampling fewer nodes when generating embeddings.

\paragraph*{Regarding Q3: Can lightweight Odin with six Transformer blocks as the text encoder also achieve excellent performance? What about efficiency?}
Limited by the semantic modeling depth of the six-layer Transformer (50\% shallower than the 12-layer BERT), its overall performance is inferior to standard Odin (12-layer Transformer) and mainstream LLM-as-Aligner baselines (e.g., Patton). The gap is more obvious especially in tasks requiring fine-grained semantic support such as retrieval and reranking — the six-layer architecture is difficult to fully extract high-level features such as entity-level correlations of text and cross-node semantic differences, leading to insufficient ability to distinguish fine-grained categories. Even so, lightweight Odin still significantly outperforms the classic cascaded fusion model PLM+sage. Moreover, the six-layer architecture reduces the computational complexity of Transformer self-attention, and the reduction in the number of TG layers (up to 2 layers) lowers the scale of neighborhood sampling. This makes lightweight Odin have a significant efficiency advantage over other models and more suitable for resource-constrained scenarios.

\paragraph*{Regarding Q4: How does Odin perform compared to mainstream pretrained language models (PLMs), graph neural networks (GNNs), and traditional GNN-LM fusion models in the four downstream tasks of node classification, link prediction, retrieval, and reranking?}
Combined with the experimental results on five datasets (Cora, CiteSeer, ArXiv2023, etc.) (Tables 8 - 11), Odin outperforms pure PLMs, pure GNNs, and fusion models in most of the four tasks. Among them, Odin's performance on the CiteSeer dataset is inferior to Patton. The reason may be that CiteSeer has a low average node degree (about 1.34) and a highly sparse graph structure, making it difficult for the GraphSAGE-minibatch multi-hop neighborhood sampling adopted by Odin to obtain a sufficiently large subgraph and resulting in insufficient capture of structural information. In contrast, Patton focuses on the modeling logic of ``local correlations of domain nodes" and has stronger adaptability to sparse graphs, thus achieving better performance on this dataset.

\section{Conclusion}
To address the long-standing challenges of semantic–structural disconnection and insufficient exploitation of high-order graph structure in text-attributed graph (TAG) representation learning, this paper introduces Odin, a hierarchical fusion framework that alternates graph aggregation with lightweight structural guidance. By aligning structure injection with the semantic depth of Transformer layers, Odin overcomes the limitations of existing LLM–GNN fusion approaches, including layer-number mismatch, rigid structure injection, and restricted structural abstraction.
Extensive experiments across five datasets and four downstream tasks demonstrate that Odin consistently outperforms strong baseline models, validating the effectiveness of its layer-aligned fusion strategy. Although the lightweight variant, Light Odin, does not reach the full model’s performance due to its smaller initialization, it still delivers substantial gains over its initial checkpoint, achieving a favorable balance between efficiency and accuracy for resource-constrained scenarios.

\bibliographystyle{sn-mathphys-ay}
\bibliography{sn-bibliography1}

\section*{Appendix A. Proof of Theorem 2}

\textbf{Proof. }We prove the theorem by showing:

(1)Odin subsumes Transformers (Odin $\geq$ Transformer).

(2)Odin subsumes GNNs (Odin $\geq$ GNN).

(3)Odin strictly surpasses both (Odin $>$ Transformer and Odin $>$ GNN).
\parindent=0pt

(1) Odin subsumes Transformers

Let $(S = \varnothing)$ , i.e., no TG layers are inserted.

In this configuration:
\parindent=15pt

•	Every layer becomes a TS layer.

•	The simple aggregation module in a TS layer can be defined as the identity map:
\begin{equation*}  
	\mathrm{Agg}_{simple}(h^{(l)}_v) = \mathbf{0} \quad \text{or} \quad \mathrm{Agg}_{simple}(h^{(l)}_v) = h^{(l)}_v.
\end{equation*}
Thus, each layer reduces to:
\begin{equation*}
	h^{(l+1)}_v = \mathrm{TransformerLayer}(h^{(l)}_v),
\end{equation*}
which is exactly the forward operator of a standard Transformer encoder layer.
Therefore, Odin can simulate any Transformer model by disabling structural injection:

$$
\text{Exp}\left( \text{Transformer} \right) \subseteq \text{Exp}\left( Odin \right) 
$$

\parindent=0pt
(2) Odin subsumes GNNs

Consider the following specialization of Odin:
\parindent=15pt

•	Let all text encoders be identity mappings:

\begin{equation*}
	\mathrm{TextEnc}(x) = x
\end{equation*}

•	Let TG layers be placed at all positions: ( S = {1,2,\dots,M} ).

•	Let the ([CLS]) representation serve as the node embedding.

•	Disable TS layers or define them as identities.

\parindent=0pt
Then the update rule becomes:

\begin{equation*}
	h^{(l+1)}_v = \sigma\left(W_1^{(l)} h^{(l)}_v + W_2^{(l)} \cdot \mathrm{mean}_{u \in N(v)} h^{(l)}_u\right)
\end{equation*}
which is exactly a GNN message-passing update such as GraphSAGE, GCN, or mean-aggregator MPNNs.

Thus, Odin can realize any function computable by message-passing GNNs:

$$
\text{Exp}\left( \text{GNN} \right) \subseteq \text{Exp}\left( Odin \right) 
$$

(3) Odin strictly surpasses both Transformers and GNNs

\parindent=15pt
To show strictness, we provide two functions:

•	one that GNNs cannot represent but Odin can,

•	one that Transformers cannot represent but Odin can.

\parindent=0pt
(3a) $Odin > \mathrm{Transformers}$

Transformers treat each node as an independent sequence without graph-induced coupling.
 
 Therefore, they cannot distinguish nodes that have identical text but different structural roles.
 
Formally, let two nodes (u, v) satisfy:

\begin{equation*}
	\mathrm{text}(u) = \mathrm{text}(v), \qquad N(u) \neq N(v)
\end{equation*}
A Transformer processes them as identical token sequences → identical outputs.
However, in Odin, the TG layer injects:
\begin{equation*}
	g^{(l)}_u = \mathrm{Agg}(N(u)), \qquad g^{(l)}_v = \mathrm{Agg}(N(v))
\end{equation*}
and if $(N(u) \neq N(v))$, then $(g^{(l)}_u \neq g^{(l)}_v)$, implying downstream representations differ.
Thus, Odin strictly distinguishes structurally distinct but textually identical nodes:

$$
\text{Exp}\left( \text{Transformer} \right) \subset \text{Exp}\left( Odin \right) 
$$

(3b) $Odin > \mathrm{GNNs}$

$\mathrm{GNNs}$ cannot distinguish textually different nodes when graph structure is symmetric or automorphic:

$$
		\exists u \neq v : N(u) \cong N(v) \quad \text{(graph automorphism)}
$$

but

$$
		\mathrm{text}(u) \neq \mathrm{text}(v)
$$

Message passing gives identical embeddings, making:
\begin{equation*}
	\mathrm{GNN}(u) = \mathrm{GNN}(v)
	\label{eq:gnn_equal}
\end{equation*}

In Odin, the text encoder produces non-equal representations:

$$
h^{(0)}_u \neq h^{(0)}_v,
$$

and no structural symmetry can collapse them through TG layers because residual-based Transformers preserve semantic separability.
Thus, Odin distinguishes textually different nodes in symmetric structures:

$$
\text{Exp}\left( \text{GNN} \right) \subset \text{Exp}\left( Odin \right) 
$$

Conclusion

Combining (1), (2), (3a), and (3b):

$$
	\mathrm{Exp}(\mathrm{Transformer}) \subset \mathrm{Exp}(Odin), \quad
$$

$$
	\mathrm{Exp}(\mathrm{GNN}) \subset \mathrm{Exp}(Odin)
$$

and there exist functions computable by Odin that neither model can express in isolation:

$$
\text{Exp}\left( \text{Transformer} \right) \cup \text{Exp}\left( \text{GNN} \right) \subset
\text{Exp}\left( Odin \right) 
$$

Thus, Odin’s expressive power strictly contains that of both Transformers and GNNs.
 $\square$

\section*{Appendix B. Category Mapping}

\label{appendix:category_mapping}

Table~\ref{tab:category_mapping} details the mapping between arXiv CS subcategories and aggregated domain groups used in our experiments, which ensures consistent classification of research topics.

\begin{table}[htbp]
	\centering
	\caption{Category Mapping for arXiv CS Domains}
	\label{tab:category_mapping}
	\begin{tabularx}{\textwidth}{X l l}  
		\toprule
		\textbf{Category Group} & \textbf{Fine-grained Category} & \textbf{Description} \\
		\midrule
		Artificial Intelligence \& Machine Learning 
		& arxiv cs ai & Artificial Intelligence \\
		& arxiv cs lg & Machine Learning \\
		& arxiv cs ne & Neural Computing \\
		\midrule
		Computer Vision \& Multimedia 
		& arxiv cs cv & Computer Vision \\
		& arxiv cs mm & Multimedia \\
		\midrule
		Natural Language Processing 
		& arxiv cs cl & Computational Linguistics \\
		& arxiv cs ir & Information Retrieval \\
		\midrule
		Systems \& Networks 
		& arxiv cs os & Operating Systems \\
		& arxiv cs dc & Distributed Computing \\
		& arxiv cs ni & Networks \\
		& arxiv cs ar & Architecture \\
		\midrule
		Data \& Databases 
		& arxiv cs db & Databases \\
		& arxiv cs ds & Data Structures \\
		& arxiv cs si & Social Information Networks \\
		\midrule
		Algorithms \& Theory 
		& arxiv cs cc & Computational Complexity \\
		& arxiv cs dm & Discrete Mathematics \\
		& arxiv cs gt & Game Theory \\
		& arxiv cs ma & Mathematical Analysis \\
		& arxiv cs ms & Mathematical Software \\
		\midrule
		Software Engineering \& Programming Languages 
		& arxiv cs se & Software Engineering \\
		& arxiv cs pl & Programming Languages \\
		& arxiv cs fl & Formal Methods \\
		& arxiv cs cr & Cryptography \& Security \\
		\midrule
		HCI \& Visualization 
		& arxiv cs hc & Human-Computer Interaction \\
		& arxiv cs cg & Computer Graphics \\
		& arxiv cs gr & Graphics \\
		\midrule
		Other 
		& arxiv cs ce & Computational Engineering \\
		& arxiv cs cy & Computers \& Society \\
		& arxiv cs dl & Digital Libraries \\
		& arxiv cs gl & General Literature \\
		& arxiv cs it & Information Theory \\
		& arxiv cs lo & Logic \\
		& arxiv cs na & Numerical Analysis \\
		& arxiv cs oh & Other \\
		& arxiv cs pf & Performance Analysis \\
		& arxiv cs ro & Robotics \\
		& arxiv cs sc & Symbolic Computation \\
		& arxiv cs sd & Speech Recognition \\
		& arxiv cs st & Statistical Theory \\
		& arxiv cs sy & Systems Theory \\
		& arxiv cs et & Educational Technology \\
		\bottomrule
	\end{tabularx}
\end{table}
Table~\ref{tab:product_category_mapping} presents the mapping between fine-grained product categories and aggregated domain groups used in our analysis.

\begin{table}[htbp]
	\centering
	\caption{Product Category Mapping}
	\label{tab:product_category_mapping}
	\begin{tabularx}{\textwidth}{X l l}
		\toprule
		\textbf{Category Group} & \textbf{Fine-grained Category} & \textbf{Description} \\
		\midrule
		Home and Living
		& Home and Kitchen & Home and Kitchen \\
		& Patio, Lawn and Garden & Patio, Lawn and Garden \\
		& Tools and Home Improvement & Tools and Home Improvement \\
		& Appliances & Appliances \\
		& Kitchen and Dining & Kitchen and Dining \\
		& Home Improvement & Home Improvement \\
		& Furniture and Décor & Furniture and Décor \\
		& Collectibles and Fine Art & Collectibles and Fine Art \\
		\midrule
		Beauty and Personal Care 
		& Beauty & Beauty \\
		& Luxury Beauty & Luxury Beauty \\
		& All Beauty & All Beauty \\
		& Cell Phones and Accessories & Cell Phones and Accessories \\
		& MP3 Players and Accessories & MP3 Players and Accessories \\
		\midrule
		Health and Baby Care
		& Health and Personal Care & Health and Personal Care \\
		& Baby Products & Baby Products \\
		& Baby & Baby \\
		\midrule
		Sports, Auto and Outdoors
		& Sports and Outdoors & Sports and Outdoors \\
		& Automotive & Automotive \\
		& GPS and Navigation & GPS and Navigation \\
		& Car Electronics & Car Electronics \\
		\midrule
		Entertainment and Media
		& Books & Books \\
		& CDs and Vinyl & CDs and Vinyl \\
		& Movies and TV & Movies and TV \\
		& Software & Software \\
		& Video Games & Video Games \\
		& Digital Music & Digital Music \\
		& Kindle Store & Kindle Store \\
		& Buy a Kindle & Buy a Kindle \\
		\midrule
		Toys, Fashion and Apparel 
		& Toys and Games & Toys and Games \\
		& Clothing, Shoes and Jewelry & Clothing, Shoes and Jewelry \\
		& Amazon Fashion & Amazon Fashion \\
		\midrule
		Electronics and Digital Gear
		& Electronics & Electronics \\
		& Computers & Computers \\
		& All Electronics & All Electronics \\
		& Camera and Photo & Camera and Photo \\
		\midrule
		Grocery and Pet Supplies
		& Grocery and Gourmet Food & Grocery and Gourmet Food \\
		& Pet Supplies & Pet Supplies \\
		\midrule
		Office and Industrial Goods 
		& Office Products & Office Products \\
		& Industrial and Scientific & Industrial and Scientific \\
		& Office and School Supplies & Office and School Supplies \\
		\midrule
		Hobbies, Gifts and Services
		& Arts, Crafts and Sewing & Arts, Crafts and Sewing \\
		& Musical Instruments & Musical Instruments \\
		& Magazine Subscriptions & Magazine Subscriptions \\
		& Purchase Circles & Purchase Circles \\
		& Gift Cards & Gift Cards \\
		\midrule
		Unclassified Labels
		& (Empty Value) & Empty Value \\
		& \#508510 & \#508510 \\
		\bottomrule
	\end{tabularx}
\end{table}
\end{document}